\documentclass[letterpaper, 10 pt, conference]{ieeeconf}  

\IEEEoverridecommandlockouts                              

\usepackage{cite}
\usepackage{booktabs}
\usepackage{amsmath,amssymb,amsfonts}
\usepackage{graphicx}
\usepackage{textcomp}
\usepackage{duckuments}
\usepackage{xcolor}
\usepackage{dsfont}
\usepackage{soul}
\usepackage{subcaption}
\usepackage[ruled,linesnumbered, vlined]{algorithm2e}
\DontPrintSemicolon
\SetCommentSty{myCommentStyle}

\usepackage{amsfonts}
\let\labelindent\relax
\usepackage{enumitem}
\usepackage{amsmath}
\usepackage{graphicx}
\usepackage{subfig}
\usepackage[margin=0.792in]{geometry}
\usepackage{bbm}
\usepackage{algpseudocode} 
\usepackage{multirow}
\usepackage{xcolor}

\newcommand\Tstrut{\rule{0pt}{2.6ex}}         
\newcommand\Bstrut{\rule[-0.9ex]{0pt}{0pt}}   
\def\BibTeX{{\rm B\kern-.05em{\sc i\kern-.025em b}\kern-.08em
    T\kern-.1667em\lower.7ex\hbox{E}\kern-.125emX}}

\title{\LARGE \bf
Integrating Active Sensing and Rearrangement Planning for Efficient Object Retrieval from Unknown, Confined, Cluttered Environments
}

\author{Junyong Kim$^+$, Hanwen Ren$^+$ and Ahmed H. Qureshi
\thanks{*This work was supported by the National Science Foundation (NSF) under award no. 2204528. $^+$Equal contribiton.}
\thanks{Junyong Kim, Hanwen Ren, and Ahmed H. Qureshi are with the Department of Computer Science, Purdue University, West Lafayette, IN, USA, 47907. Email {\tt\small$\{$kim3722, ren221, ahqureshi$\}@$purdue.edu}}%
}

\begin{document}

\maketitle
\thispagestyle{empty}
\pagestyle{empty}

\begin{abstract}
Retrieving target objects from unknown, confined spaces remains a challenging task that requires integrated, task-driven active sensing and rearrangement planning. Previous approaches have independently addressed active sensing and rearrangement planning, limiting their practicality in real-world scenarios. This paper presents a new, integrated heuristic-based active sensing and Monte-Carlo Tree Search (MCTS)-based retrieval planning approach. These components provide feedback to one another to actively sense critical, unobserved areas suitable for the retrieval planner to plan a sequence for relocating path-blocking obstacles and a collision-free trajectory for retrieving the target object. We demonstrate the effectiveness of our approach using a robot arm equipped with an in-hand camera in both simulated and real-world confined, cluttered scenarios. Our framework is compared against various state-of-the-art methods. The results indicate that our proposed approach outperforms baseline methods by a significant margin in terms of the success rate, the object rearrangement planning time consumption and the number of planning trials before successfully retrieving the target. Videos can be found at https://youtu.be/tea7I-3RtV0.

\end{abstract}

\section{INTRODUCTION}
The ability to retrieve target objects from an unknown, confined, cluttered environment is a crucial skill for robots aiming to assist people in their daily lives.  For example, in homes, robots will often need to retrieve objects from tight spaces like refrigerators and cabinets. In hospitals, they might need to fetch medical instruments for staff, while in factories, they may fetch tools for human workers. Despite various applications, solving the problem of retrieving objects from unknown, confined spaces remains a challenge for robots due to the following reasons. 
 
First, it is challenging to perceive the confined scenarios, as it's difficult and impractical to place external perception sensors inside them. The robot must rely on its onboard sensors to understand the environment. Second, retrieving the target object may require moving non-target objects in a partially-observed environment to allow the robot to reach and retrieve the desired object without collisions. Third, the unknown environments may be large, making perception time-consuming. Additionally, these environments are often rearranged, which means that the robot needs to perceive the environment every time it needs to retrieve an object. 

The previous research in robot active sensing and object retrieval only addresses some of the challenges mentioned above and assumes that the rest have already been solved. For example, existing object retrieval methods may deal with rearranging objects and assume that the environment is known. Meanwhile, active sensing methods focus on maximizing the scene coverage without considering the underlying object retrieval tasks.

In this paper, we propose an integrated active sensing and retrieval planning approach for target object retrieval from unknown, cluttered, and confined environments. Our method begins to solve the object retrieval problem once the target is detected via active sensing, i.e., with only limited partial scene observation. Given the partial observation containing the target object, our novel approach presents an integrated plan for rearranging non-target objects and actively sensing unobserved areas that are crucial for clearing the robot pathway to reach and retrieve the target object without collision. 
The key elements and core contributions of our new approach are summarized as follows:
\begin{itemize}[leftmargin=*]
    \item A novel task-driven active sensing approach that detects the target object in an unknown environment and leverages feedback from the retrieval planner to selectively sense the remaining unobserved regions that will aid in retrieving the target object. 

    \item A Monte-Carlo Tree Search (MCTS) based retrieval planner that selects the non-target objects blocking the robot's pathway to the target object and plans their relocation regions while minimizing the overall robot motion distance. Our MCTS-based planner also selects the unobserved regions of the environment for active sensing that are currently being treated as obstacles but have the potential to allow the successful inference of a retrieval plan. 

    \item A unified, integrated framework that tightly combines active sensing and retrieval planning to solve object retrieval problems in unknown, cluttered, confined environments.
\end{itemize}
We demonstrate our approach to solving various object retrieval tasks using a robot arm with an in-hand camera in both simulation and real-world confined, cluttered scenarios. We compare our approach against various state-of-the-art methods. Our results indicate that our approach outperforms baseline methods in retrieving the desired objects.

\section{Related Work}
The prior work views active scene sensing and object retrieval problems independently, and to the best of our knowledge, their synergy has never been studied. 

The active scene sensing method aims to plan a sequence of viewpoints that maximizes the scene coverage or leads to the detection of any given target. These methods can be categorized into classical and learning-based approaches. The former \cite{vasquez2014view, bircher2016receding} design frontier- and sampling-based methods to generate the next-best-view which maximizes a pre-defined utility function value. The latter \cite{zeng2020pc, mendoza2020supervised, ren2023robot} learns from either expert demonstrations or trial-and-error-based data to infer a policy predicting the next best viewpoint for scene sensing. Despite advancements, current sensing methods do not utilize feedback from the retrieval planner to guide sensing toward relevant areas for the retrieval task.

The problem of object retrieval has also been studied independently of active sensing by assuming the complete scene can be captured by a single static camera. Thus, the majority of work in this domain focuses on cluttered tabletop environments \cite{9636230, zhong2022soft, rouillard2019autonomous, kim2019retrieving}. For instance, Huang et al. \cite{huang2021visual} proposed a method called Visual Foresight Trees (VFT) for moving surrounding objects efficiently with non-prehensile robot actions such as pushing to reach the target object. In a similar vein, Michael et al. \cite{8794143} expanded the robot action space by using multiple grippers allowing suction, pushing, and grasping to retrieve a target object obstructed by novel obstacles. Some of the existing work also considers partially observed scenarios. For example,  \cite{xiao2019online, 9636230} formulated the problem as a Partially Observable Markov Decision Process (POMDP) to address uncertainty during the search process, leading to higher success rates.

As tabletop environments are not always representative of real-world scenarios, object retrieval tasks are also studied in confined spaces, which imposes complex collision avoidance constraints on robot motion for retrieving a target object. The current methods for devising a non-target object rearrangement plan to reach and retrieve the target object mainly fall into two categories: algorithmic search-based methods \cite{9561282, nam2019planning, bejjani2021occlusion, nam2021fast, 10341865, 9197485} and data-driven learning-based methods \cite{9332268, ren2024neuralrearrangementplanningobject, tang2023selective}. The algorithmic method often leads to better generalization with a completeness guarantee, whereas learning-based methods may produce infeasible paths when generalizing to novel scenarios. These methods face challenges in perceiving confined environments. Some rely on a single viewpoint image \cite{9332268}, while others \cite{ren2024neuralrearrangementplanningobject} use dense scene sensing to maximize scene coverage before initiating the retrieval planning. However, single-viewpoint images often result in poor performance, especially in cluttered scenarios. Whereas, dense sensing may lead to redundant viewpoints and still not guarantee the coverage of critical regions essential for robot navigation in confined environments.

The problem of object retrieval planning can also be viewed as a subset of object rearrangement planning problems \cite{ kang2023object, krontiris2015dealing, chen2023optimal, qureshi2021nerp}. In the retrieval problem, the non-target objects need to be rearranged to make the way for the robot to reach and retrieve the target object. Whereas in rearrangement planning, the methods mainly deal with the task of reaching target object rearrangement from the initial object arrangements \cite{qureshi2021nerp, 9036915, 8793946, 7487583, wang2022lazy}. These tasks are also mostly studied in the tabletop settings with a few exceptions of recent Monte Carlo Tree Search (MCTS)-based methods \cite{chaslot2008monte}. These methods \cite{kang2023object,ren2024multistagemontecarlotree} consider known confined environments and use MCTS policy to devise a rearrangement strategy. Our work is also inspired by the MCTS-based planners due to their capability in finding high-quality solutions in complex scenarios by balancing exploration and exploitation. However, in our work, we integrate MCTS with active sensing to handle object retrieval tasks, which require rearrangement planners to empty the space that the robot may occupy in reaching and retrieving the target object.  
\section{Methods}
This section formally presents our integrated active sensing and retrieval planning approach. We begin by introducing the necessary notations, followed by our method details.

Let an unknown confined environment be denoted as $\mathcal{S} \in \mathbb{R}^4$ with dimensions $d_x$, $d_y$, and $d_z$ along $x$, $y$ and $z$ axes. The fourth dimension of $\mathcal{S}$ indicates the observed and unobserved parts. We denote the observed and unobserved parts of the scene as $\mathcal{S}_o$ and $\mathcal{S}_{uo} = \mathcal{S} \backslash \mathcal{S}_o$. In scene $\mathcal{S}$, 3D pixel locations associated with $\mathcal{S}_{o}$ are marked with indicator 1, while the remaining parts are denoted with indicator 0 to signify unobserved areas. At first, the environment is entirely unobserved. Inside the scene, there exists a target object $o^*$. Besides $o^*$, a group of $m$ sizeable objects $\{o_0,...,o_m\}$ also exist in the scene $\mathcal{S}$, making the scene cluttered and object retrieval impossible without rearrangement of the non-target objects. In our setting, a robot is a manipulator with an in-hand camera attached to its end effector. The configuration space of the robot is denoted as $\mathcal{Q} \in \mathbb{R}^d$, where $d$ represents the robot's degree of freedom. 

The active sensing action is a 6D viewpoint $\boldsymbol{v} \in \mathbb{R}^6$ that places the robot's in-hand camera at the pose given by $v$. A coverage metric $\phi(\mathcal{S}_x(\boldsymbol{v}))$ is defined based on a pin-hole camera model to measure the pixel that will be marked as observed in an arbitrary scene representation $\mathcal{S}_x$ based on viewpoint $\boldsymbol{v}$, i.e.,\vspace{-0.1in}
\begin{equation}\vspace{-0.1in}
    \phi(\mathcal{S}_x(\boldsymbol{v})) = \frac{\sum_i^{d'_x}\sum_j^{d'_y}\sum_k^{d'_z} \mathcal{S}_x(i,j,k) \neq 0}{d'_x\cdot d'_y\cdot d'_z},
\end{equation}
where $d'_x$, $d'_y$, and $d'_z$ are the dimensions of arbitrary scene $\mathcal{S}_x$. The rearrangement action $\boldsymbol{a}=\{(a^0_{pick}, a^0_{place}),\cdots, (a^k_{pick}, a^k_{place})\}$ is a sequence of pick and place locations of non-target objects, i.e., $O=\{o^0, \cdots, o^k\}$, in the observed scene $\mathcal{S}_o$. 
\subsection{Multi-objective Active Sensing}
This section introduces our active sensing functions that are employed by our retrieval planner to selectively sense an unknown environment. 
\subsubsection{Scene Coverage Sensing}
In this phase, we aim to detect the given target object in an unknown environment. Since we lack prior knowledge of the target object's location, the most effective strategy is to maximize scene coverage during active sensing until the target object is observed. Hence, our objective is to actively sense an unknown environment in order to detect a specific target object using a minimum number of viewpoints, i.e.,\vspace{-0.05in}
\begin{equation}\vspace{-0.05in}
     \max_\phi \min_T \sum_{t=0}^{T} \phi(\mathcal{S}(\boldsymbol{v}_{t}))
     \textrm{ s.t. } \mathds{1}(o^* \in \mathcal{S}_o) = 1
\end{equation}
To achieve the above objective, we propose to utilize existing active sensing approaches that maximize the scene coverage with a minimum number of viewpoints. Specifically, we use the ViewPointFormer (VPFormer) \cite{ren2023robot}, which is an imitation learning-based transformer neural network approach that iteratively generates viewpoints based on the given scene representation. 
The outcome of this phase is an initial scene representation $\mathcal{S}_I=(\mathcal{S}_o,\mathcal{S}_{uo})$, where $\mathcal{S}_o$ contains the target object $o^*$.

\subsubsection{Region-specific Sensing}
This module presents an active sensing policy, denoted as $\pi_{rs}$, that observes a specific region of a partially observed environment $\mathcal{S}_I$. Let the desired unobserved region be denoted as $\mathcal{\bar{S}}_{uo} \subset \mathcal{S}_{uo}$. The objective of our region-specific sensing is to maximize the coverage of $\mathcal{\bar{S}}_{uo}$ in a minimum number of viewpoints, i.e.,\vspace{-0.05in}
\begin{equation}\vspace{-0.05in}
     \max_\phi \min_{T'} \sum_{t=0}^{T'} \phi(\mathcal{\bar{S}}_{uo}(\boldsymbol{v}_{t}))
\end{equation}
To solve the above objective function, we propose a heuristic-based viewpoint policy that is both recursive and greedy. The policy $\pi_{rs}$ selects the centroid of a particular unobserved region $\mathcal{\bar{S}}_{uo}$ as a focus for a viewpoint $\boldsymbol{v}$. It then uses the random viewpoint pose sampling and selects the best viewpoint using a pin-hole camera model that produces the most intersecting rays with $\mathcal{\bar{S}}_{uo}$ while the region's centroid being the focus. This viewpoint often results in complete coverage of $\mathcal{\bar{S}}_{uo}$ in confined spaces. However, if $\mathcal{\bar{S}}_{uo}$ is not entirely observed, we divide the remaining unobserved regions into clusters and recursively observe each cluster using the aforementioned greedy approach. Finally, this sensing module returns the new scene representation, which also contains the observation of $\mathcal{\bar{S}}_{uo}$. Note that we propose a non-learning-based approach for region-specific sensing, which in our settings has proven to be more effective in terms of time efficiency and coverage. 
\subsection{Integrated Sensing and Object Retrieval Planning }
The following sections present our object retrieval planer, which utilizes the aforementioned active sensing functions to generate object retrieval plans for execution. 
\subsubsection{Retrieval Reachability Sensing and Planning}
This function comprises the reachability-based planning and sensing stage. During the reachability planning phase, the function, denoted as $f_{orp}$, takes the initial scene representation $\mathcal{S}_I$ and target object $o^*$ and outputs a robot path, $\sigma^{o^*}$, and its swept volume, $V_\sigma^{o^*}$, to reach and retrieve that target object. Additionally, it also computes the $\sigma^{o^*}$-blocking non-target objects $O=\{o_0,\cdots,o_k\}$ and their robot reaching paths and swept volumes i.e., 
\begin{equation}\label{orp}\small
\sigma^{o^*}, V_\sigma^{o^*}, O, \{\sigma^0_{pick},V^0, \cdots, \sigma^k_{pick},V^k \}    \gets f_{orp}(\mathcal{S}_I,o^*)
\end{equation}
Our approach starts by generating a set of target object grasping poses $\{G\} = \{g_1,...,g_w\}$ using a pre-trained Contact-GraspNet \cite{sundermeyer2021contact}. For each grasp $g_i$, we leverage a motion planning tool with smoothing \cite{844730} to generate the robot motion sequence for reaching and retrieving the target object without considering other objects in the environment. Let this path be denoted as $\sigma^{o^*}_i \subseteq \mathcal{Q}$ under grasp $g_i$. Among all potential grasps and paths, we choose the one with the smallest swept volume to increase the probability of finding feasible object retrieval plans. Let this selected path and its swept volume be denoted as $\sigma^{o^*}$ and $V_\sigma^{o^*}$, respectively. Note that we can analytically compute the swept volume using robot forward kinematics. Additionally, the robot's target object reaching and retrieval paths and swept volumes will be different since, during retrieval, the robot will also be holding the target object. Next, the non-target objects $O$ are selected that intersect with $V_\sigma^{o^*}$. 
For each path-blocking object $o_j \in O$, we use a motion planning tool to determine a picking path, $\sigma^j_{pick}$, to reach that object. Let that path and their corresponding swept volumes be denoted as $\{\sigma^0_{pick},V^0, \cdots, \sigma^k_{pick},V^k \}$. 

During the reachability plan-based active sensing stage, this module employs our region-specific sensing policy $\pi_{rs}$ to observe the swept volumes $\{V_\sigma^{o^*}, V^0, \cdots, V^k\}$ in case any of those regions belong to unobserved areas $\mathcal{S}_{uo}$, leading to an updated scene representation $\mathcal{S}'_I$

\subsubsection{Retrieval Rearrangement Sensing and Planning}
In this stage, our method utilizes the outcomes of the reachability planning and sensing phase (Eq. \ref{orp}) and devises an integrated active sensing and retrieval plan. Our \textbf{retrieval policy}, $\pi_r$ is based on Monte Carlo Tree Search (MCTS), which aims to relocate path-blocking non-target objects $O$ with minimum move distance to unoccupied regions in the observed scene $\mathcal{S}'_I$ such that those regions are outside robot swept volume $V_\sigma^{o^*}$. Let's represent the observed scene configuration after applying a robot's non-target objects' relocation action sequence $\boldsymbol{a}$ as $\mathcal{S}_o(\boldsymbol{a})$. Let a function, $d(\cdot)$, measure the Euclidean distance between the given inputs. Then, our MCTS objective is to find the optimal action sequence $\boldsymbol{a}$ that minimizes the non-target objects relocation effort and successfully retrieves the target object, i.e.,\vspace{-0.1in}
\begin{equation*}\vspace{-0.05in}\small
    \text{argmin}_{\boldsymbol{a}} \mathbb{E}_{\boldsymbol{a} \sim \pi_r}  \bigg[\sum_{i=0}^k d(a^i_{pick}, a^i_{place})\bigg] \:
    \textrm{s.t.} \: \:  V_\sigma^{o^*} \cap \mathcal{S}_o(\boldsymbol{a}) = \emptyset
\end{equation*}
To achieve the above objective, we design an Object Retrieval MCTS (OR-MCTS) which uses the MCTS\cite{chaslot2010monte} and SS-MCTS \cite{ren2024multistagemontecarlotree} in its backbone. The SS-MCTS is originally designed for reaching a desired target arrangement of a given scene. Our OR-MCTS extends SS-MCTS to solve the retrieval tasks in partially observed confined environments. In our OR-MCTS, each tree node represents one specific scene configuration. In addition, the parent node and the child node are linked by a single pick-n-place object relocation action. During the search process, the method first selects an existing tree node based on a tuned Upper Confidence Bound (UCB) \cite{browne2012survey} value that favors those that have the lowest object relocation distance and those that are not adequately exploited. Then, the tree expands by proposing several closest feasible regions outside the target swept volume, $V_\sigma^{o^*}$, for all the accessible non-target objects in $O$. Additionally, unlike SS-MCTS, we also give our planner the capability to relocate objects inside the target swept volume $V_\sigma^{o^*}$ when the current tree node cannot grow further. We observe that this feature is crucial for solving retrieval tasks in partially observed confined spaces. For objects that move into the target swept volume, the planner ensures that such action opens up reachability to previously unreachable objects in $O$ to make it one step closer to solving the task. For unreachable objects, the method figures out the object dependency relation and clears the way by moving other objects first. All the tree nodes go through a simulation process to numerically evaluate how close they are to solving the problem. Next, the evaluation value back-propagates through the tree to update the UCB values for all tree nodes. This search process keeps running until, at a certain node, the target swept volume is entirely empty. Backward traversal from that node leads to retrieval plan $\boldsymbol{a}$.
In case our OR-MCTS succeeds in reaching the task objects, it returns a retrieval plan $\boldsymbol{a}$ for execution. However, in case of failure, it provides feedback to region-specific sensing policy, $\pi_{rs}$ to observe the selected unobserved regions. Recall that our method treats unobserved regions as obstacles, which can lead to failures as our OR-MCTS cannot expand trees where there is a lack of available free space for relocating non-target objects. Thus, our method leverages the failed OR-MCTS plan to guide active sensing.  

The \textbf{OR-MCTS feedback-based active sensing} starts by extracting the scene configuration $\mathcal{S}_o'$ from the most promising child node of the latest failed OR-MCTS. The promising child node is selected as one that has the lowest remaining non-target objects to be relocated. In the case of multiple candidate child nodes, we select the one with the largest reward value. From $\mathcal{S}_o'$, the region growing method \cite{6203897} is applied to link the adjacent unobserved spaces, which results in a set of unobserved clusters $\{\mathcal{S}_{uo}^1,...,\mathcal{S}_{uo}^n\}$. Then, for each cluster $\mathcal{S}_{uo}^j$, our method calculates the sum of feasible relocation regions it opens up for the remaining objects inside the swept volume $V_\sigma^{o^*}$. During the feasible location region counting process, the algorithm explicitly considers the complete robot motion to evaluate the unobserved clusters for selection. The cluster providing the largest reachable relocation region is selected. We use our policy $\pi_{rs}$ to observe that selected region. After active sensing, the $\pi_{rs}$ returns the new scene representation $\mathcal{S}'_I$, and the above OR-MCTS function is re-executed to find a feasible retrieval place. The integrated retrieval planning and sensing are repeated until a retrieval plan is successfully inferred or a time limit is reached, leading to a failure. 
\vspace{-0.1in}\subsection{Execution Algorithm}
Algorithm \ref{alg:full_algo} outlines our integrated sensing and retrieval planning method. The proposed Multi-objective Active Sensing (MAS) consists of three stages. (1) \textbf{Initial Active Sensing (IAS):} Detects the target object. (2) \textbf{swept volume Active Sensing (SAS):} Actively senses swept volumes related to reaching and retrieving the target object and reaching path-blocking non-target objects. (3) \textbf{Feedback-based Active Sensing (FAS):} Performs sensing based on feedback from OR-MCTS. The procedure begins with IAS that detects the target object $o^*$ (Line 2). Next, the reachability planning module (Line 3) identifies the robot's path, $\sigma^{o^*}$, and its swept volume $V^{o^*}_\sigma$ to reach and retrieve the target object $o^*$. Additionally, it detects non-target objects $O'$ for relocation that resides inside $V^{o^*}_\sigma$, and the robot's swept volumes to reach those objects. These swept volumes are sensed via SAS, leading to $\mathcal{S}'_I$ (Line 4). Observing these swept volumes is crucial, as otherwise, OR-MCTS will treat them as obstacles that may lead to failures. Next, the OR-MCTS plans the first rearrangement sequence $\boldsymbol{a}$ to move objects $O'$ outside the sweep area $V^{o^*}_\sigma$ (Line 5). If OR-MCTS succeeds, the plan $\boldsymbol{a}$ is executed; otherwise, the algorithm leverages the FAS module to view the next best region that can potentially lead to a successful target retrieval conditioned on the most promising tree node in the the failed OR-MCTS attempt (Lines 7-8). When new feedback-based observation is obtained, OR-MCTS is deployed again (Line 9). The process repeats until the scene is entirely observed or our planner infers the solution (line 6).\vspace{-0.1in}
\begin{algorithm}[hbt!]
\caption{OR-MCTS via MAS}\label{alg:full_algo}
$\mathcal{S}_I = \emptyset, O = \emptyset, o^*=\text{None}$ \Comment{Initilization}\;
$o^*, \mathcal{S}_I, O \leftarrow \text{IAS}(\mathcal{S}_I)$ \;
$\sigma^{o^*},V_\sigma^{o^*}, O, \{\sigma^i_{pick},V^i\}^k_{i=0} \leftarrow f_{orp}(\mathcal{S}_I,o^*)$ \Comment{Reachability Planning}\;
$\mathcal{S}'_I, O'\leftarrow \text{SAS}(V_\sigma^{o^*}, \{V^i\}^k_{i=0})$\;
$\boldsymbol{a}, \mathcal{T} \leftarrow \text{OR-MCTS}(V_\sigma^{o^*}, O', \mathcal{S}'_I)$ \Comment{first trial}\;
\While{$\boldsymbol{a} = \emptyset$ \text{and} $\mathcal{S}_i \neq \mathcal{S}_{env}$}{
    $\mathcal{S}'_{uo} \leftarrow \text{Cluster\_UnObserved\_Region}(\mathcal{T}, \mathcal{S}'_I)$\;
    $\mathcal{S}'_I \leftarrow \text{FAS}(\mathcal{S}'_{uo})$\; 
    $\boldsymbol{a}, \mathcal{T} \leftarrow \text{OR-MCTS}(V_\sigma^{o^*}, O, \mathcal{S}'_I)$ \Comment{new trial}\;
}

\Return $\boldsymbol{a}$
\end{algorithm}\vspace{-0.15in}
\begin{figure}[h]\vspace{-0.15in}
\centering
\begin{subfigure}{0.35\textwidth}
\centering
\includegraphics[trim={0 0 0 15cm},clip, width=1.0\linewidth]{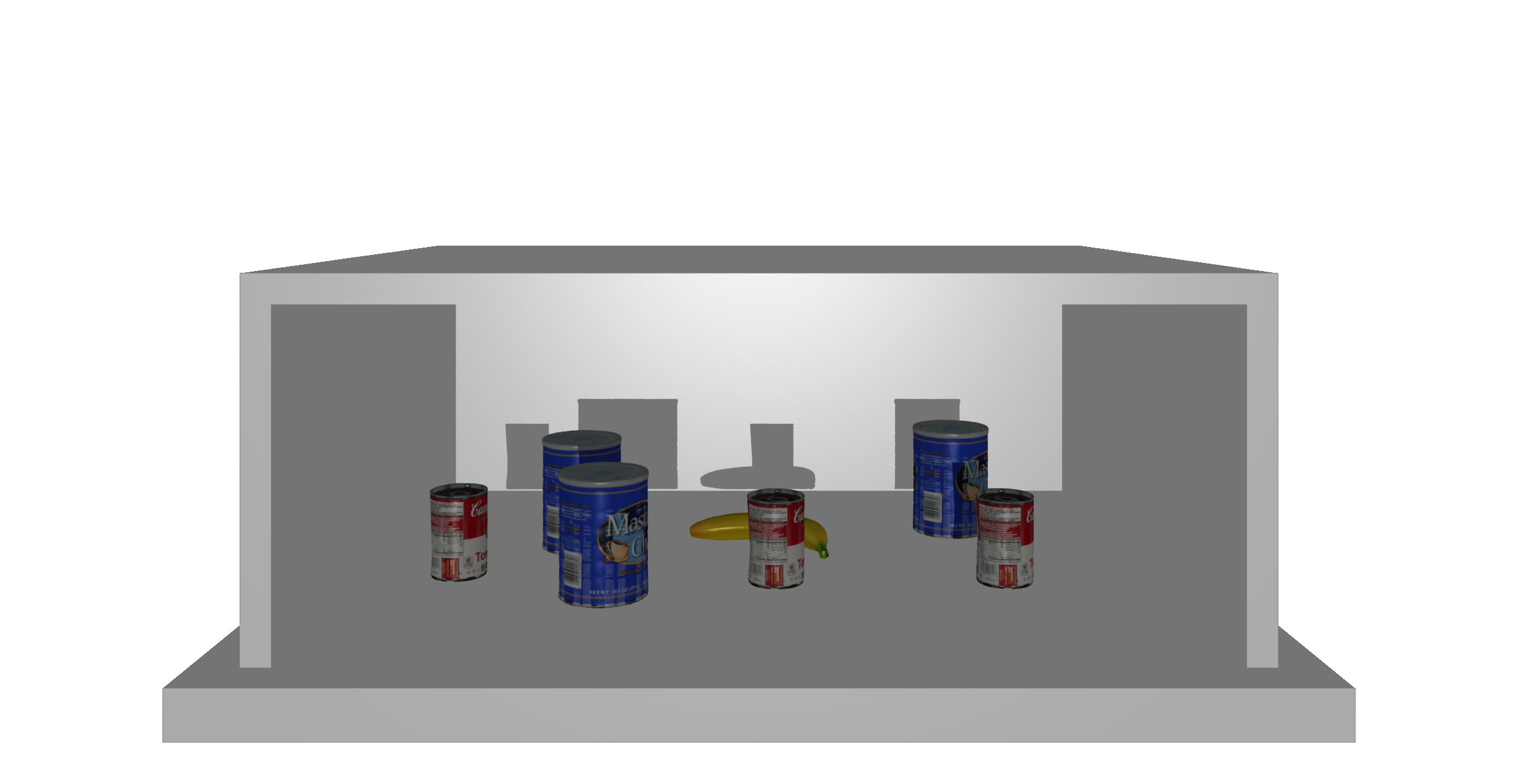}\vspace{-0.1in}
\caption{Small scene: target object (Banana)}
\label{fig:capparatus}
\end{subfigure}
\begin{subfigure}{0.35\textwidth}
\centering
\includegraphics[trim={0cm 6cm 0 0cm},clip,width=1.0\linewidth]{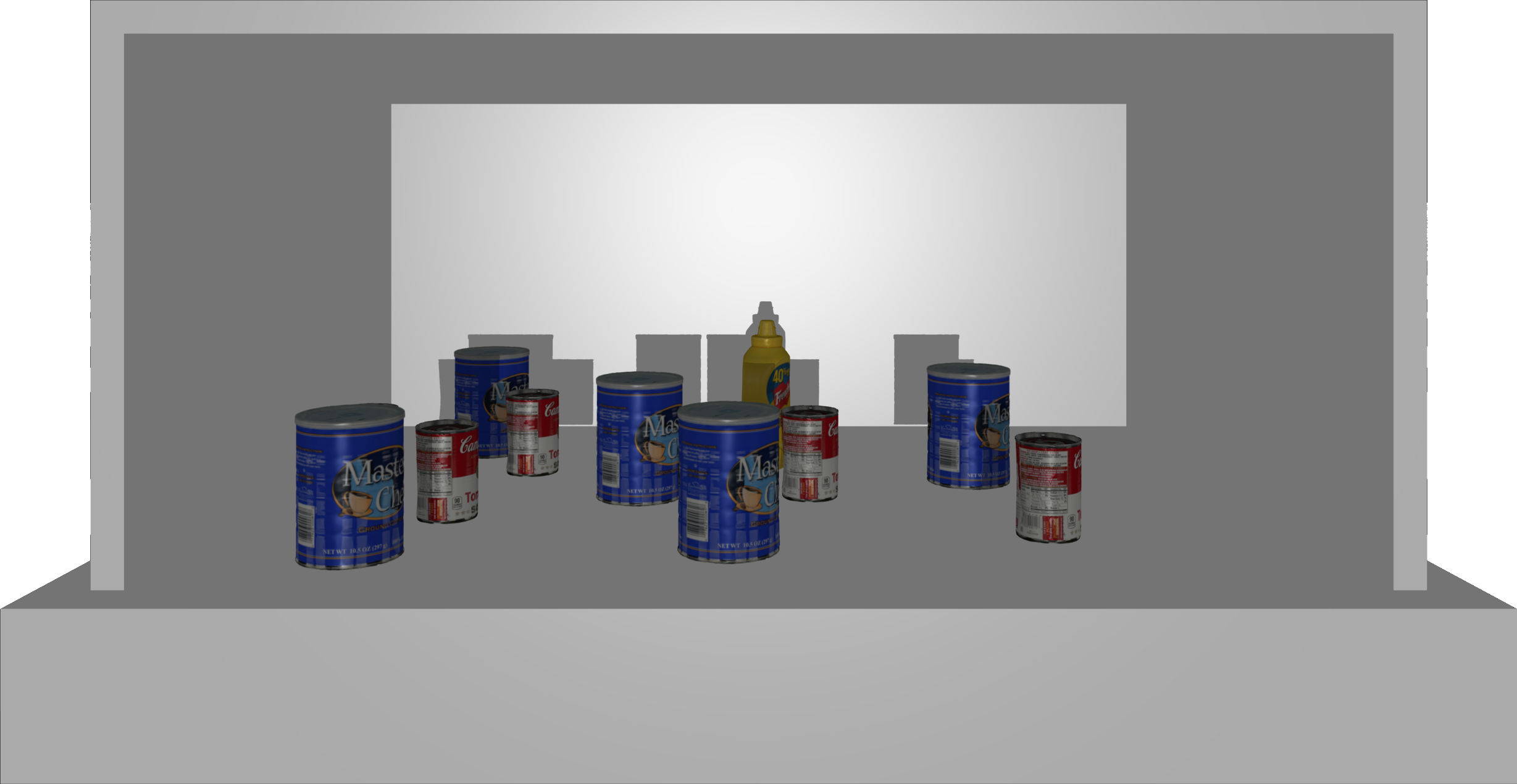}
\caption{Large scene: target object (mustard bottle)}
\label{fig:cdiagram}
\end{subfigure}

\caption{ Depiction of cluttered example testing environments.}
\label{testscenes}\vspace{-0.2in}
\end{figure}
\section{Experiments and Discussions}
\begin{figure*}[t]
\centering
\begin{subfigure}{0.24\textwidth}
\centering
\includegraphics[trim={1.5cm 0.2cm 0.65cm  0cm},clip, width=1.0\linewidth]{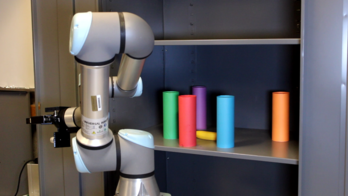}
\caption{Initial scene configuration}
\end{subfigure}
\begin{subfigure}{0.24\textwidth}
\centering
\includegraphics[trim={1.5cm 0.2cm 0.65cm 0cm},clip,width=1.0\linewidth]{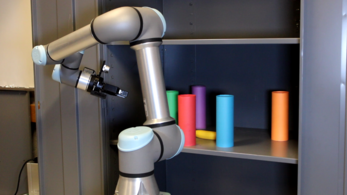}
\caption{IAS to detect target}
\end{subfigure}
\begin{subfigure}{0.24\textwidth}
\centering
\includegraphics[trim={1.5cm 0.2cm 0.65cm 0cm},clip,width=1.0\linewidth]{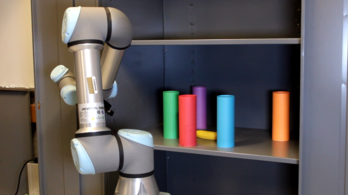}
\caption{SAS to observe swept volumes}
\end{subfigure}
\begin{subfigure}{0.24\textwidth}
\centering
\includegraphics[trim={1.5cm 0.2cm 0.65cm 0cm},clip,width=1.0\linewidth]{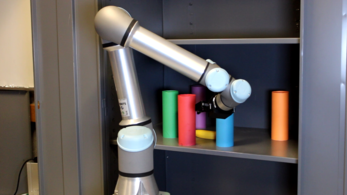}
\caption{FAS via OR-MCTS feedback}
\end{subfigure}
\begin{subfigure}{0.24\textwidth}
\centering
\includegraphics[trim={1.5cm 0.2cm 0.65cm 0cm},clip,width=1.0\linewidth]{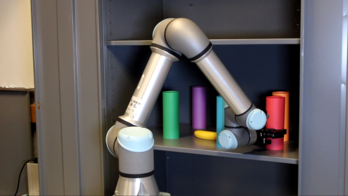}
\caption{Relocating red cylinder}
\end{subfigure}
\begin{subfigure}{0.24\textwidth}
\centering
\includegraphics[trim={1.5cm 0.2cm 0.65cm 0cm},clip,width=1.0\linewidth]{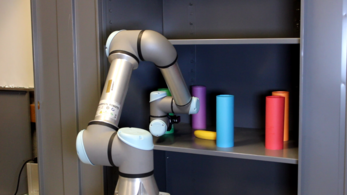}
\caption{Relocating green cylinder}
\end{subfigure}
\begin{subfigure}{0.24\textwidth}
\centering
\includegraphics[trim={1.5cm 0.2cm 0.65cm 0cm},clip,width=1.0\linewidth]{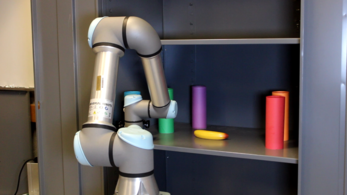}
\caption{Relocating blue cylinder}
\end{subfigure}
\begin{subfigure}{0.24\textwidth}
\centering
\includegraphics[trim={1.5cm 0.2cm 0.65cm 0cm}, clip, width=1.0\linewidth]{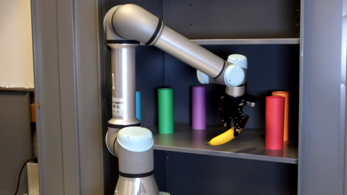}
\caption{Target successfully retrieved}
\end{subfigure}
\caption{Our method, MAS+OR-MCTS, solves the object retrieval task in the confined, real-world cabinet environment shown in (a) where the target object banana is surrounded by 5 sizeable cylindrical movable obstacles. The approach starts by actively sensing the environment to (b) detect the target object, (c) observe the robot swept volumes for it to be able to begin running OR-MCTS for target retrieval, and (d) other critical areas based on the FAS + OR-MCTS. Once a feasible object rearrangement plan is found, the robot executes it (e-g), resulting in successful retrieval of the target object (h).}
\label{real}\vspace{-0.25in}
\end{figure*}

In this section, we present our experiments, their analysis, and related discussion. We randomly generated 100 testing scenarios varying their sizes between small and large scenes as shown in Fig. \ref{testscenes}. In each generated scene, we randomly position an arbitrary target object surrounded by randomly placed five to eight non-target objects to make them realistically cluttered. In these environments, we evaluate our active sensing and object retrieval planning approaches against various baseline methods. Additionally, we also evaluate our method in real-world confined cabinet settings. In all of the above scenarios, we use a 6-DOF UR5e robot arm with an in-hand camera (Fig. \ref{real}). To compare all methods, we employ the following metrics. Additionally, all experiments are performed on a system with 3.0 GHz $\times$ AMD Ryzen 9 5900HS, 16 GB RAM, and GeForce RTX 3070 GPU.
\begin{itemize}[leftmargin=*]
    \item \textbf{Success rate (SR) ($\%$)}: 
    The experiment is considered a success only if a method can generate a feasible object retrieval plan before the scene is entirely observed.
    
    \item \textbf{Time (sec)}: This metric tracks the time consumption for the planner to find a feasible plan.

    \item \textbf{Number of attempts (\# attempts)}: It is the total number of times the retrieval planner fails to find a plan within a time budget of 30 seconds and reverts back to leverage FAS for new observations.

    \item \textbf{Number of objects moved (\# objects moved)}: It represents the number of prehensile non-target objects relocation actions required before target retrieval.
    
    
    \item \textbf{Relocation distance (cm)}: It denotes the sum of the Euclidean distance of the path-blocking objects' movements.

    \item \textbf{Number of viewpoints (\# viewpoints)}: It describes the total number of observations made until termination.
\end{itemize}

\begin{table}[h]\vspace{-0.1in}

  \begin{center}
    \scalebox{0.82}{
    \begin{tabular}{c c c c c}
      \toprule
       Method & SR (\%) $\uparrow$ \Tstrut\Bstrut & time (sec) $\downarrow$ & \# viewpoints $\downarrow$ & \# attempts $\downarrow$\\
      \midrule
     MAS (Ours) \Tstrut\Bstrut& \textbf{95} \Tstrut\Bstrut& 10.20 ± 9.71 \Tstrut\Bstrut& 3.71 ± 0.93 & 1.41 ± 0.69 \\
     
     IAS+FAS  (Ours) \Tstrut\Bstrut& 92 \Tstrut\Bstrut& 9.71 ± 9.57 \Tstrut\Bstrut& 3.74 ± 1.07 & 3.97 ± 1.89 \\
     
     IAS+SAS \Tstrut\Bstrut& 61 \Tstrut\Bstrut& 18.08 ± 12.74 \Tstrut\Bstrut& 3.29 ± 0.86 & 1.00 ± 0.00 \\
     
    DIAS \cite{ren2023robot} \Tstrut\Bstrut& 27 \Tstrut\Bstrut& 24.25 ± 11.51 \Tstrut\Bstrut& 4.42 ± 1.76 & 1.00 ± 0.00 \\
    
    IAS \Tstrut\Bstrut& 4 \Tstrut\Bstrut& 38.75 ± 24.08 \Tstrut\Bstrut& 1.26 ± 0.52 & 1.00 ± 0.00 \\
      \bottomrule
    \end{tabular}}
    \caption{Comparison shows integrated sensing and retrieval planning (MAS and IAS+FAS) methods demonstrate significantly better overall performance.}
    \label{tab:table1}
  \end{center}
\vspace{-0.36in}
\end{table}
\begin{table*}[!ht]
  \begin{center}
    \begin{tabular}{c c c c  c c}
      \toprule
       Method& SR (\%) $\uparrow$ \Tstrut\Bstrut & time (sec) $\downarrow$& \# attempts $\downarrow$ & \# objects moved $\downarrow$ & relocation distance (cm) $\downarrow$ \\
      \midrule
     MAS + OR-MCTS (Ours) \Tstrut\Bstrut& \textbf{95} \Tstrut\Bstrut& \textbf{10.20 ± 9.71} \Tstrut\Bstrut&  \textbf{1.41 ± 0.69} \Tstrut\Bstrut& 3.83 ± 0.95 \Tstrut\Bstrut& 57.55 ± 22.61\\
     
     MAS + SS-MCTS \cite{ren2024multistagemontecarlotree} \Tstrut\Bstrut& 78 \Tstrut\Bstrut& 14.90 ± 10.89 \Tstrut\Bstrut& 1.66 ± 0.96 \Tstrut\Bstrut& 3.46 ± 0.65 \Tstrut\Bstrut& \textbf{44.78 ± 16.41}\\
     
     MAS + TSAD \cite{10341865} \Tstrut\Bstrut& 79 \Tstrut\Bstrut& 14.54 ± 10.94 \Tstrut\Bstrut& 1.56 ± 1.10 \Tstrut\Bstrut& 4.61 ± 2.66 \Tstrut\Bstrut& 137.71 ± 88.99 \\
     
     MAS + WTL \cite{9197485} \Tstrut\Bstrut& 59 \Tstrut\Bstrut& 18.58 ± 11.84 \Tstrut\Bstrut& 2.02 ± 1.32 \Tstrut\Bstrut& \textbf{3.41 ± 0.76} \Tstrut\Bstrut& 152.22 ± 47.55\\
      \bottomrule
    \end{tabular}
    \caption{ Retrieval planning comparison: All of the methods use our MAS for scene perception. Our approach outperforms others by a significant margin. The relative higher numbers of objects moved and relocation distance are attributed to our method's ability to solve harder cases requiring various object relocation, in which other methods have failed.}
    \label{tab:table2}
  \end{center}
\vspace{-0.15in}
\end{table*}

\subsection{Effectiveness of integrated active sensing and planning}
This section aims to evaluate the significance of integrating active sensing and retrieval planning in the simulation environments. To assess the effectiveness of each stage in our MAS, we create the following baselines. All of them use our OR-MCTS as the underlying object retrieval planner.
\begin{itemize}[leftmargin=*]
    \item \textbf{MAS:} Integrated pipeline with IAS, SAS, and FAS.
    \item \textbf{IAS + SAS:} This baseline excludes FAS and relies on IAS to detect the target and SAS to sense the swept volumes. 
    \item \textbf{IAS + FAS:} The method excludes SAS and relies on IAS to detect the target and FAS to provide necessary scene coverage based on the feedback from the OR-MCTS.
    \item  \textbf{IAS:} In this baseline, IAS detects the target object, after which no further active sensing strategy is performed. 
    \item \textbf{Dense Initial Active Sensing (DIAS):} This method densely sense the entire scene using a recent approach \cite{ren2023robot} to maximize the coverage percentage until at least 95\%.
\end{itemize}\vspace{-0.02in}

A comparison between our approach and the above-mentioned baseline methods is presented in Table \ref{tab:table1}. 
It is evident that methods using FAS, such as MAS and IAS+FAS, outperform other methods significantly in terms of SR and time. On the other hand, MAS outperforms IAS+FAS in terms of the number of attempts, highlighting the necessity of sensing the robot swept volumes (SAS) for the retrieval planner to succeed; otherwise, it relies on FAS to observe those regions, which is inefficient. In addition, IAS+SAS underperforms compared to MAS and IAS+FAS, as it cannot capture the critical regions that can potentially serve as the non-target object relocation regions, which emphasizes the importance of our FAS. For the DIAS baseline, it densely observes the whole scene until a decent scene coverage is reached. However, as shown in Table \ref{tab:table1}, the performance of DIAS is significantly worse compared to our approach. Similar to IAS+SAS, DSAI often fails capture the crucial tiny regions for relocation as they are usually occluded. Finally, IAS exhibited poor performance due to insufficient scene coverage by just detecting the target. The reason why IAS uses fewer viewpoints than other baselines is due to fact that it stops sensing after the target object was detected. In summary, from the above results, it can be concluded that MAS provides the overall best performance, and the feedback from the retrieval planner to guide active sensing (FAS) is critical for target retrieval.\vspace{-0.05in}
\subsection{Evaluation of Object Retrieval Planning:} 
This section evaluates our OR-MCTS planner against the following baseline methods in the same 100 simulation environments as the active sensing tests. All baselines are augmented with our MAS for perception as they originally assume a known environment. 
\begin{itemize}[leftmargin=*]
    \item \textbf{SS-MCTS\cite{ren2024multistagemontecarlotree}:} SS-MCTS shares a similar structure as our OR-MCTS. The main difference is that it does not allow the usage of the target retrieval swept volume.
    \item \textbf{Tree Search in Approaching Direction (TSAD)\cite{10341865}:} TSAD utilizes a sequence of MCTS to move one object each time. During the search, it proposes new relocation regions using a random policy. 
    Each MCTS terminates when the selected node removed at least one object from the swept volume for target retrieval. The search ends when the swept volume is entirely cleared and returns the corresponding plan.
    \item \textbf{Where To Relocate (WTL)\cite{9197485}:} WTL divides the area into valid and invalid regions with invalid ones being the location where an object should not be placed, such as robot swept volumes and unobserved areas. 
    The algorithm selects moves that minimize the creation of new invalid regions. In the case of no valid region remains, it randomly places a feasible object into an invalid region.
\end{itemize}

Table \ref{tab:table2} summarizes our results. Our OR-MCTS method significantly outperforms all other methods in terms of SR, time, and the number of attempts. The relatively higher relocation distance and objects moved in our method are associated with its superior performance, as it can handle more challenging cases that require long horizon of relocation actions. The relatively poor performance of SS-MCTS suggests that prohibiting the placement of non-target objects in the robot swept volume for target object retrieval limits MCTS's performance in cluttered environments. The TSAD's subpar performance is due the usage of a random policy for selecting the relocation region, which greatly increases computation time and object relocation distance. Finally, WTL exhibits the lowest performance, indicating their approach is not suitable for cluttered, confined scenarios. In all, these results reveal that our OR-MCTS is ideal for solving retrieval tasks in confined spaces. Additionally, integrating our MAS into SS-MCTS, TSAD, and WTL gives them the capability to accomplish some tasks in unknown environments. These methods would have suffered severely if they had relied solely on dense sensing approaches such as DSAI, as evident from results in Table \ref{tab:table1}. \vspace{-0.1in}
\subsection{Real Robot Experiments}
Next, we deploy our method in the real-world cabinet setting shown in Fig. \ref{real}. In the cabinet environment, we generated five scenes by randomly placing five to six sizeable non-target objects surrounding the target object. On average real-world results of our method are similar in statistics to simulation results, indicating sim to real generalization of our approach. The planning time for the instance depicted in Fig. \ref{real} is 9.38 s. It successfully retrieves the target with an object relocation distance of 89.80 cm after a single call to FAS. However, sometimes failure can occur regarding the segmentation algorithm failing to detect the objects under our cluttered and low-lighting environments. Please refer to our supplementary material for the demonstration videos.\vspace{-0.05in}
\section{Conclusions and Future Work}
This paper presents the first integrated active sensing and object retrieval planning approach. Our retrieval planner is based on the Monte-Carlo Tree Search, which generates a rearrangement sequence to relocate non-target objects that block the retrieval path for the target object. Our retrieval planner also provides feedback to our active sensing model to observe the unknown regions of the environment that are critical for solving the retrieval tasks. The results demonstrate that our approach outperforms baselines by a significant margin and highlight that the integration of active sensing and planning is essential for solving object retrieval problems in confined, cluttered environments. In our future work, we aim to computationally enhance our retrieval planner with a data-driven scoring function that would prevent the expansion of less promising MCTS nodes, leading to even faster solutions to the object retrieval tasks.

\bibliographystyle{IEEEtran}
\bibliography{root}

\end{document}